\documentclass[dvipsnames]{article}
\usepackage{iclr2020_conference,times}
\pdfoutput=1

\usepackage{booktabs} 
\usepackage{hyperref}
\usepackage{url}
\usepackage{graphicx}
\usepackage{floatrow}
\newfloatcommand{capbtabbox}{table}[][\FBwidth]
\usepackage{blindtext}

\usepackage[T1]{fontenc}
\usepackage[utf8]{inputenc}
\usepackage[font=small,labelfont=bf]{caption}

\usepackage{siunitx}
\usepackage{booktabs, makecell}

\usepackage{capt-of}
\usepackage{varwidth}
\newsavebox\tmpbox
\title{Comparing recurrent and convolutional neural networks for predicting wave propagation}

\author{Stathi Fotiadis, Eduardo Pignatelli,  Anil A. Bharath \\
Department of Bioengineering\\ Imperial College London\\
\And
Mario Lino Valencia, Chris D. Cantwell \\
Department of Aeronautics\\Imperial College London\\
\AND
Amos Storkey\\
Institute for Adaptive and Neural Computation\\
The University of Edinburgh
}

\newif\ifincludecomment
\includecommenttrue 
\ifincludecomment
  
\else
  \NewEnviron{guidance}{}{}
\fi
\usepackage[colorinlistoftodos,textsize=tiny, textwidth=18mm]{todonotes}
\ifincludecomment
\newcommand{\maybecomment}[1]{\todo[color=olive!40]{#1}} 
\newcommand{\maybetohere}[1]{\todo[color=red!40]{#1}} 
\newcommand{\maybedelete}[1]{\todo[color=blue!40]{#1}} 

\else
  \newcommand{\maybecomment}[1]{}
  
\newcommand{\maybedelete}[1]{} 
\fi
\newcommand{\amostohere}[1]{{\color{black}\maybetohere{AMOS HERE}}}

\iclrfinalcopy 
\begin{document}

\maketitle

\begin{abstract}
Dynamical systems can be modelled by partial differential equations and a need for their numerical solution appears in many areas of science and engineering. In this work, we investigate the performance of recurrent and convolutional deep neural network architectures to predict the propagation of surface waves governed by the Saint-Venant equations. We improve on the long-term prediction over previous methods while keeping the inference time at a fraction of numerical simulations. We also show that convolutional networks perform at least as well as recurrent networks in this task. Finally, we assess the generalisation capability of each network by extrapolating for longer times and in different physical settings.\footnote{Code and data available at 
\url{github.com/stathius/wave_propagation}
}

\end{abstract}

\section{Introduction}

Many physical systems in science and engineering are described by partial differential equations (PDEs). This study investigates the performance of recurrent and convolutional deep neural networks to model such phenomena. Accurately predicting the evolution of such systems is usually done through numerical simulations, a task that requires significant computational resources. Simulations usually need extensive tuning and need to be re-run from scratch even for small variations in the parameters.
With their potential to learn hierarchical representations, deep learning techniques have emerged as an alternative to numerical solvers, by offering a desirable balance between accuracy and computational cost \citep{Carleo2019MachineSciences}.

Here, we focus on the modelling of surface wave propagation governed by the Saint-Venant (SV) equations. This phenomenon offers a good test-bed for controlled analyses on two-dimensional sequence prediction of PDEs for several reasons. First, in contrast to some physical systems, such as fluid flow, the evolution of the real system is unlikely to enter chaotic regimes. From a representation learning point of view, this makes model training and assessment relatively straightforward.  Despite this, the SV equations are strongly related to the Navier-Stokes equations, widely used in computational fluids.  Further, computational modelling of surface waves is used in seismology, computer animation, in predictions of surface runoff from rainfall -- a critical aspect of the water cycle \citep{moussa2000approximationsaint} -- and flood modelling \citep{ersoy2017saintvenant}. 

This study provides three contributions. First, we identify three relevant architectures for spatiotemporal prediction. Two of these architectures lead to improved  accuracy in long-term prediction over previous attempts \citep{Sorteberg2019ApproximatingNetworks} while keeping the inference time orders of magnitude smaller than typical solvers.
Secondly, our comparison between recurrent and purely convolutional models indicates that both can be equally effective in spatiotemporal prediction of SV PDEs. This is in alignment with the findings of \cite{bai2018empirical} that demonstrates that convolutional models are as effective as recurrent models in one-dimensional sequence modelling. %
Finally, we evaluate the generalisation of the models in situations not seen during training and indicate their shortcomings.

\begin{minipage}[b]{0.5\textwidth}
\vspace{0pt}
\centering
\includegraphics[width=\linewidth]{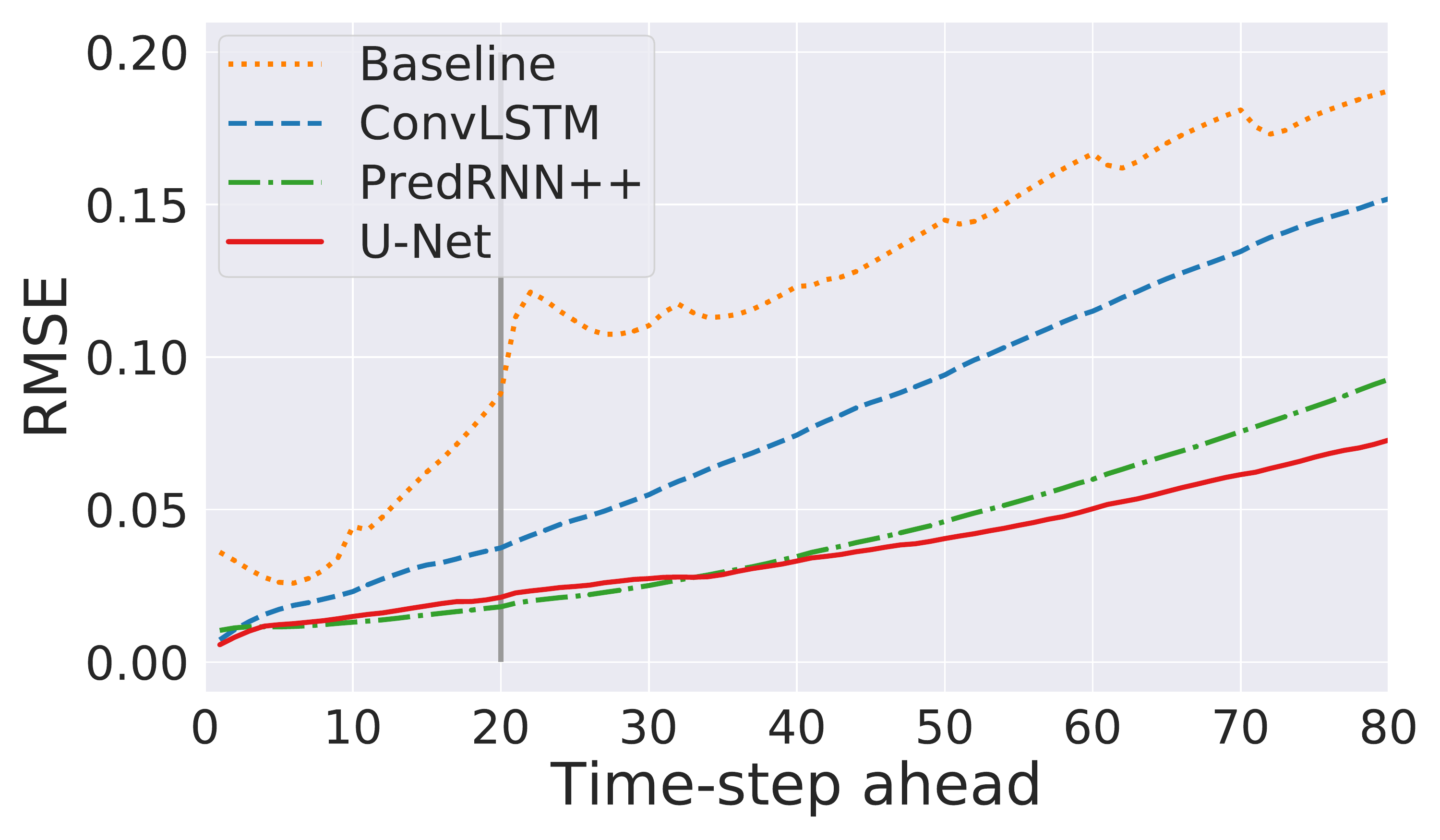} 
\captionof{figure}{\textbf{Comparing the long-term prediction of the four models on the test set.} The Causal-LSTM and the U-Net significantly outperform the baseline LSTM model. The vertical line indicates the training horizon.}
\label{fig:res:model_comp}
\end{minipage}
\hfill
\begin{minipage}[b]{0.49\textwidth}
\vspace{0pt}
\centering

\begin{tabular}{lrr}
\toprule
Time-step ahead &    20 &    80 \\
\midrule
LSTM (baseline)        & 0.08 $\pm$ 0.00 & 0.19 $\pm$ 0.03 \\
ConvLSTM    & 0.05 $\pm$ 0.00& 0.15 $\pm$ 0.01 \\
PredRNN++ & \textbf{0.02} $\pm$ \textbf{0.01} & 0.09 $\pm$ 0.01\\
U-Net       & \textbf{0.02} $\pm$ \textbf{0.00} & \textbf{0.07} $\pm$ \textbf{0.01}\\
\bottomrule
\end{tabular}
\vspace{1cm}

\captionof{table}{\textbf{Root Mean Square Error (RMSE) comparison of model prediction at specific time-steps ahead.} Best accuracy in bold. The standard errors are across 4 runs with different initialisation.}
\label{tab:res:model_comp}
\end{minipage}

\begin{figure}[ht]
\centering
\begin{minipage}[l]{.49\textwidth}
  \includegraphics[width=\linewidth]{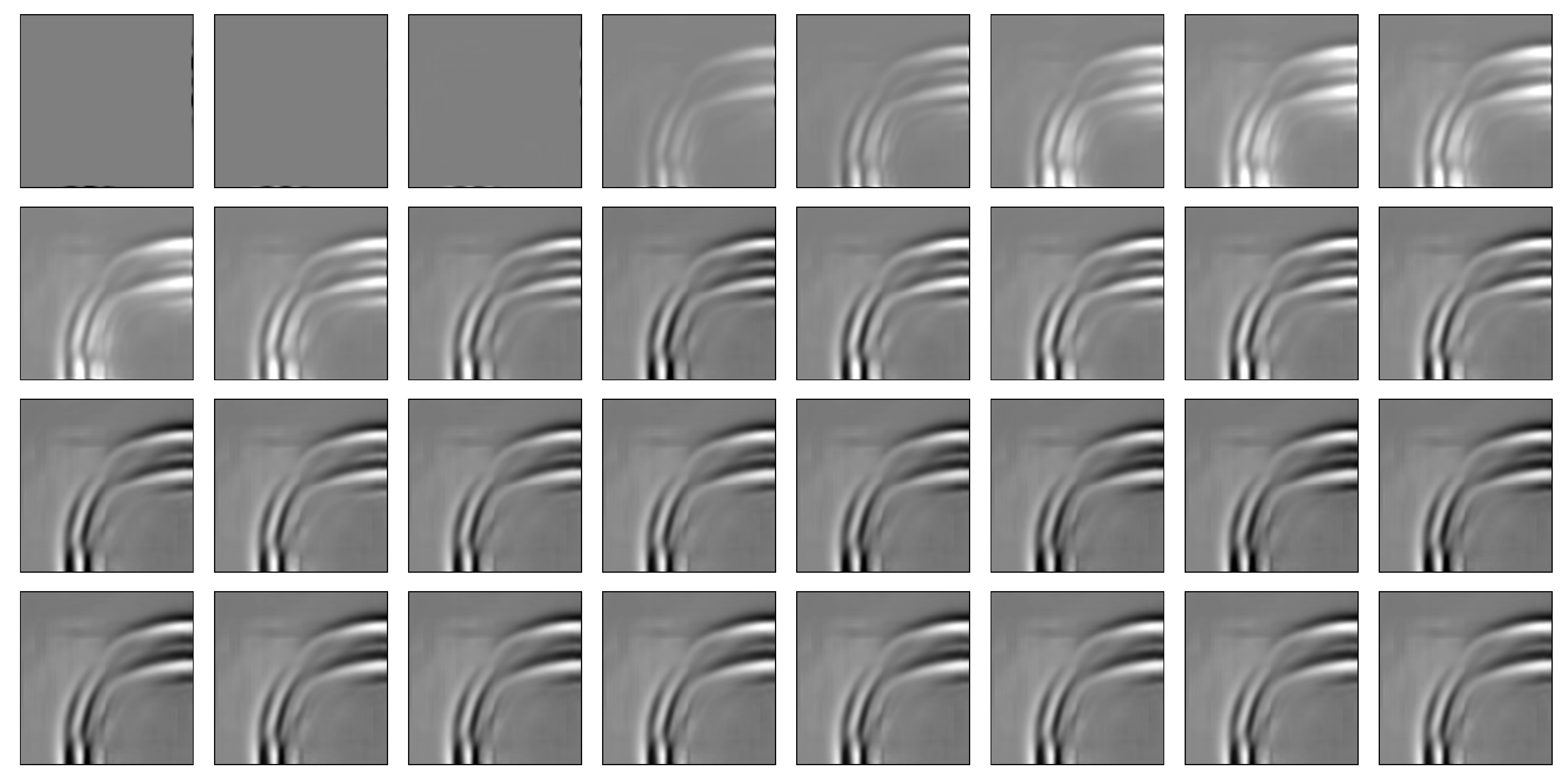} 
\end{minipage}%
\hfill
\begin{minipage}[r]{.49\textwidth}
  \includegraphics[width=\linewidth]{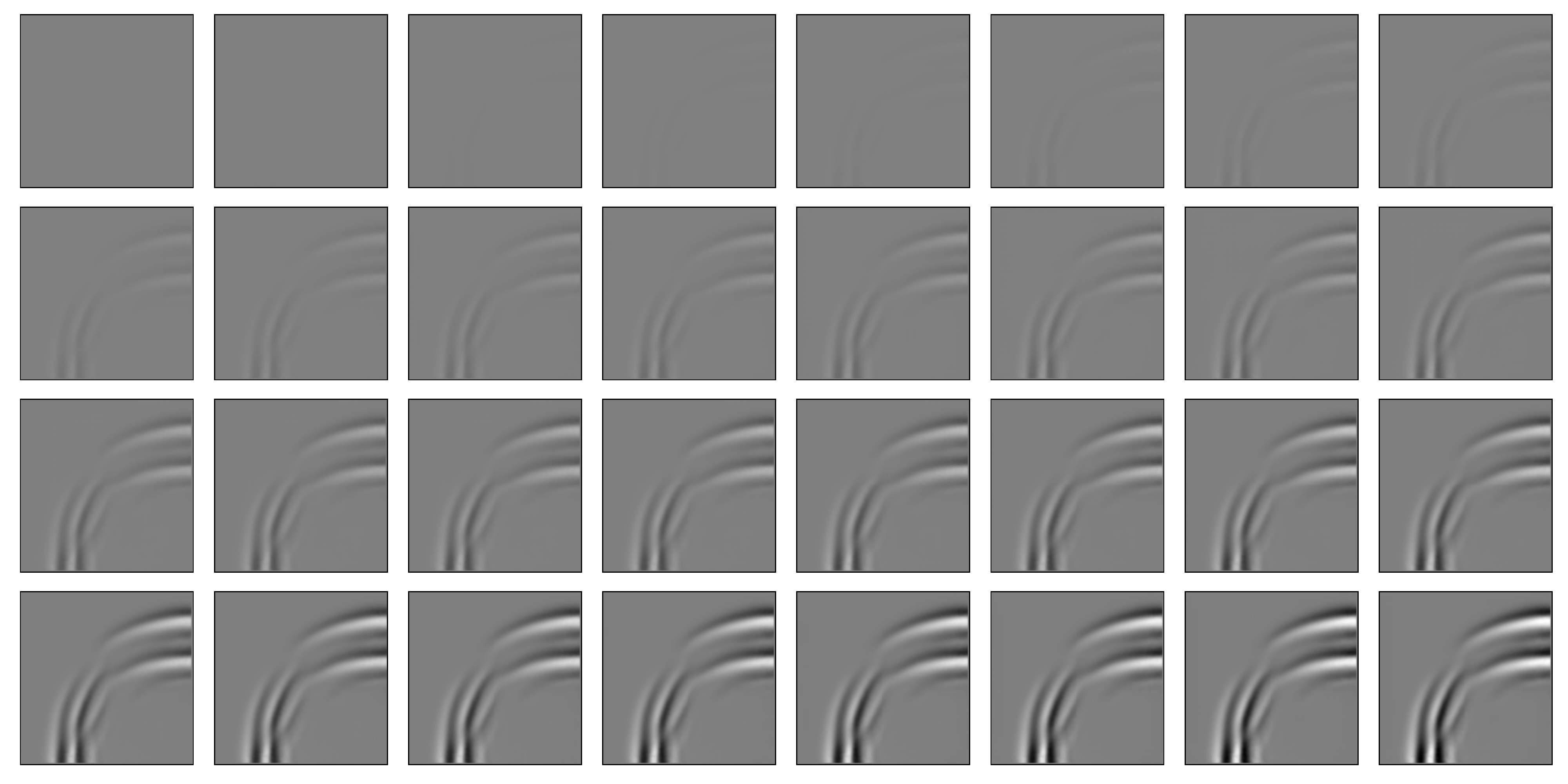}
\end{minipage}
  \caption{\textbf{Cumulative reconstruction of the output from the feature maps of the pre-last layer of the \mbox{U-Net} (top) and PredRNN++ (bottom).} Prediction corresponds to the 80th time-step ahead. We ordered the feature maps by the absolute value of their weight, from the most important to the least. The PredRNN++ gradually builds up its prediction. The U-Net works differently: the first few feature maps put emphasis on the boundary conditions. Then some of the feature maps focus on the peaks (white colour) and some others on the troughs. All combined build the final prediction. 
  }
  \label{add:fig:reconstruct}
\vskip -2mm
\end{figure}

\vskip -5mm
\section{Related work}
\label{sec:related_work}
\vskip -2mm
Deep learning methods have been proposed for spatiotemporal forecasting in various fields including the solution of PDEs. Recurrent neural networks have been proven a good fit for the task, due to their innate ability to capture temporal correlations. \cite{srivastava2015unsupervised} use a convolutional encoder-decoder architecture where an LSTM module is used to propagate the latent space to the future. Variations of this technique have been successfully applied to the long-term prediction of physical systems, such as sliding objects \citep{Ehrhardt2017LearningPredictor} and wave propagation \citep{Sorteberg2019ApproximatingNetworks}.
Convolutional LSTMs (ConvLSTM) use convolutions inside the LSTM cell to complement the temporal state with spatial information. 
Whilst initially proposed for  precipitation nowcasting, ConvLSTMs were also found successful for video prediction \citep{Shi2015ConvolutionalNowcasting}. \cite{Wang2018PredRNN++:Learning} proposed the PredRNN++, featuring spatial memory that traverses the stacked cells in the network and improves the accuracy of short-term prediction over ConvLSTMs.

Feed-forward models have, also, been used in spatiotemporal forecasting. \cite{mathieu2015deep} used a CNN to encode video frames in a latent space and extrapolated the latent vectors to the future. \cite{tompson2017accelerating} employed CNNs to speed up the projection step in fluid flow simulations. U-Net has been used for optical flow estimation in videos \citep{DosovitskiyFlowNet:Networks} as well as in physical systems, such as sea temperature predictions \citep{bezenac2017deep} and accelerating the simulation of the Navier-Stokes equations \citep{Thuerey2018DeepFlows}. While both recurrent and convolutional models have been successfully applied for the prediction of PDEs, there is a paucity of studies comparing the two categories from a representation learning point of view.

Other architectures for spatiotemporal prediction include Generative Adversarial Networks, for fluid simulations \citep{Kim2018DeepSimulations} and Graph Networks for wind-farm power estimation \citep{Park2019Physics-inducedEstimation}. There is also a growing body of research on physics-inspired networks for solving PDEs \citep{Raissi2017PhysicsEquations, Perdikaris2019ModelingModels}.


\begin{minipage}[t]{0.49\textwidth}
\vspace{0pt}
\centering
\includegraphics[width=\linewidth]{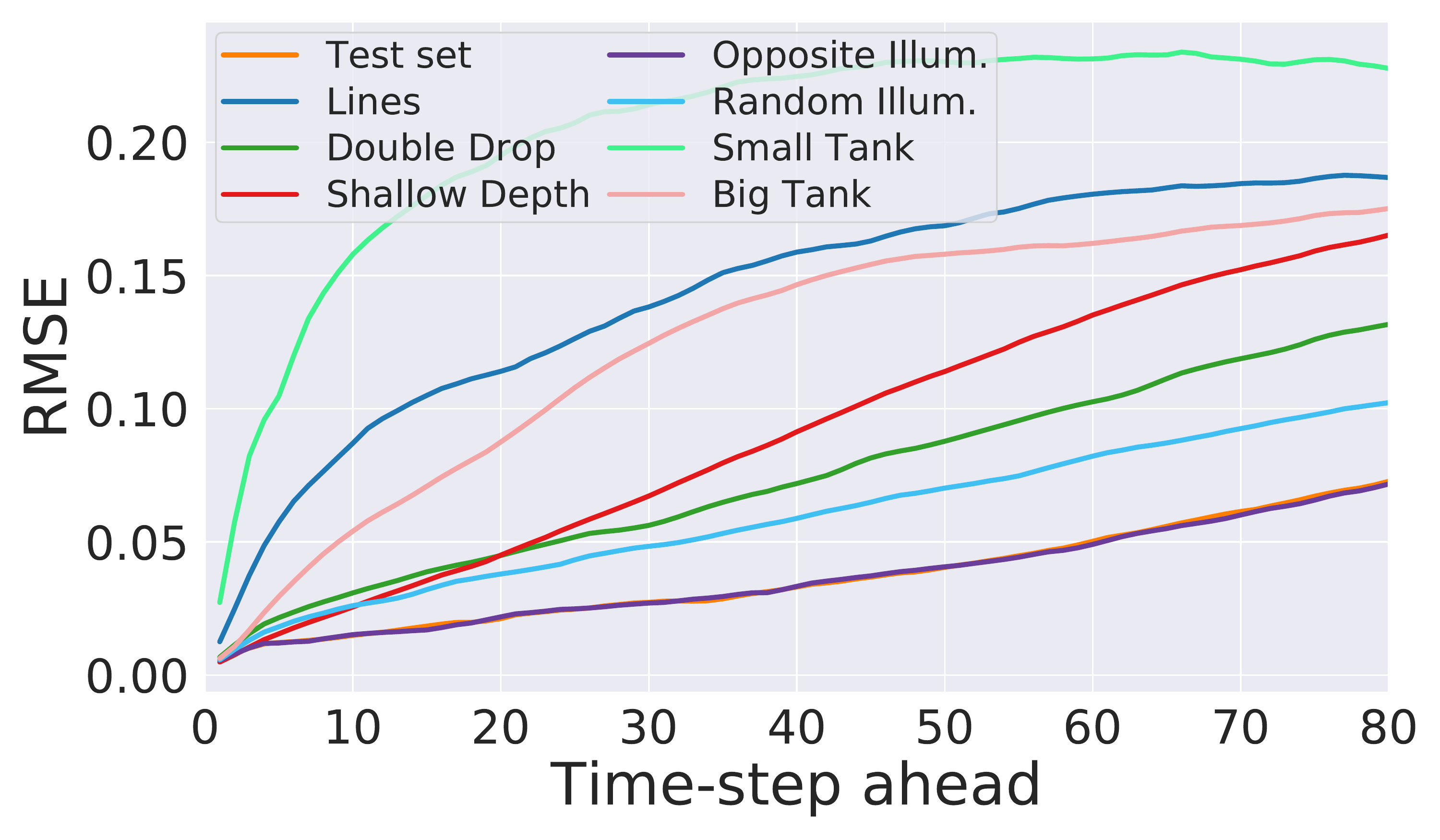} 
\captionof{figure}{\textbf{Generalisation in different physical settings for the U-Net.} The network copes well with changes to illumination and even with two drop but cannot predict well linear waves or different tank size.}
\label{fig:res:gen}
\end{minipage}
\hfill
\begin{minipage}[t]{0.49\textwidth}
\vspace{0.1cm}
\centering
\small
\begin{tabular}{lrrrr}
\toprule
Time-step ahead &    20 &    40 &    60 &    80 \\
\midrule
Test set        & \textbf{0.02} & \textbf{0.03} & \textbf{0.05} & \textbf{0.07} \\
Opposite Illum. &\textbf{ 0.02} &\textbf{ 0.03} & \textbf{0.05 }&\textbf{ 0.07} \\
Random Illum.   & 0.4 & 0.06 & 0.08 & 0.10 \\
Double Drop     & 0.04 & 0.07 & 0.10 & 0.13 \\
Lines           & 0.11 & 0.16 & 0.18 & 0.19 \\
Shallow Depth   & 0.04 & 0.09 & 0.13 & 0.16 \\
Big Tank        & 0.08 & 0.14 & 0.16 & 0.17 \\
Small Tank      & 0.19 & 0.22 & 0.23 & 0.23 \\
\bottomrule
\end{tabular}
\vspace{0.3cm}
\captionof{table}{\textbf{RMSE of U-Net across datasets at specific points in time.} Performance varies across different physical settings. The model is invariant to an orthogonal phase shift in illumination.}
\label{tab:res:gen}
\end{minipage}

\section{Evaluated models}

Four different models are assessed in this work. Three of them are recurrent (LSTM, ConvLSTM, PredRNN++) and one is feed-forward (U-Net). A detailed description of all the implementations can be found in Section \ref{app:sec:mod} of the Appendix. The LSTM model was specifically developed for wave propagation prediction \citep{Sorteberg2019ApproximatingNetworks} and serves as a baseline on which we sought improvement. It is composed of a convolutional encoder and decoder with three LSTMs in the middle. The LSTM modules use the vector output of the encoder as an inner representation and propagate it forward in time. Each LSTM propagates a different part of the sequence (see Appendix).

The other models were selected on the basis of their applicability to relevant tasks. ConvLSTM and PredRNN++ have been empirically shown to perform well at short-term spatiotemporal predictions. The rationale for using them in long-term prediction is that the underlying physics of wave propagation do not change. If a model learns a good  representation of short-term dynamics, then the error accumulation should remain low long-term. Both models use convolutions inside the recurrent cell to create a synergy between spatial and temporal modelling. Additionally, PredRNN++ employs a spatial memory that traverses the vertical stack to increase short-term accuracy.

The feed-forward model is based on the U-Net architecture used in spatiotemporal prediction. For example, it has been used to infer optical flow \citep{fischer2015flownet} , motion fields \citep{bezenac2017deep} and velocity fields \citep{Thuerey2018DeepFlows}. In contrast, we train the network end-to-end and conditional on its own predictions; the latter 
shifts the focus from short-term to long-term accuracy.

\section{Results}\label{sec:results}

\subsection{Long term prediction: Extrapolation in time}

We evaluated how well the models  extrapolate in time. Given ground-truth simulations of 100 frames in length, we tested the model predictions up to 80 steps, much more than the maximum of 20 frame sequences that the models are trained upon.
The RMSE at each time step is calculated as an average over all the test sequences.  Results show that the baseline LSTM gives the worst performance. The RMSE error reaches 0.10 after only 21 frames while the error sharply raises after frame 10 (Figure \ref{fig:res:model_comp}). A probable cause is the usage of three distinct LSTMs, which require more data to train upon. The ConvLSTM offers an improvement: it reaches 0.1 RMSE after only 53 frames. The error trend is also very gradual, almost linear. An even greater improvement comes from the PredRNN++, which provides a very low error over the whole prediction range. Its maximum error at frame 80 is 0.091, substantially lower than the LSTM (0.186) and the ConvLSTM (0.150) (Table \ref{tab:res:model_comp}). This confirms the findings of \cite{wang2018predrnn++}, that PredRNN++ is more efficient than ConvLSTM. U-Net is on par with PredRNN++ until frame 34, but has better long-term prediction, reaching 0.071 RMSE at frame 80 vs 0.091 of the PredRNN++. The U-Net decreases the RMSE by $62\%$ compared to the baseline. It is also the faster model, providing a $240\times$ speed-up over the numerical solver that we used (Table \ref{app:tab:speed} in Appendix).

Qualitatively, it appears that the PredRNN++ propagates its internal representation one step at a time while the U-Net predicts multiple frames in one pass. How the output is reconstructed in the last layer is indicative of the differences (Figure  \ref{add:fig:reconstruct}).

\begin{figure}[t]
\centering
\begin{minipage}[c]{.4\textwidth}
  \centering
  \includegraphics[width=\linewidth]{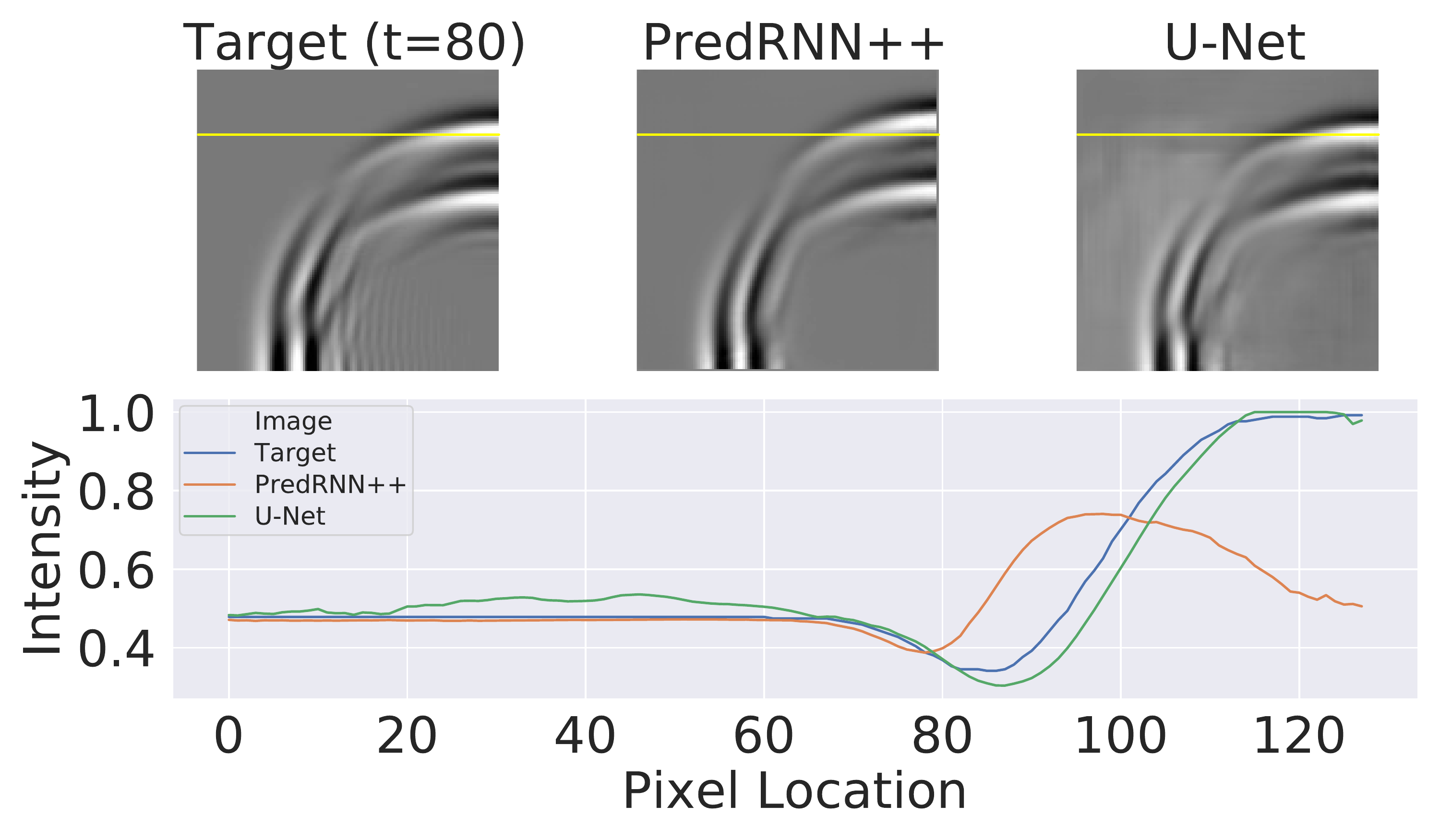} 
\end{minipage}%
\hfill
\begin{minipage}[c]{.58\textwidth}
  \centering
  \includegraphics[width=\linewidth]{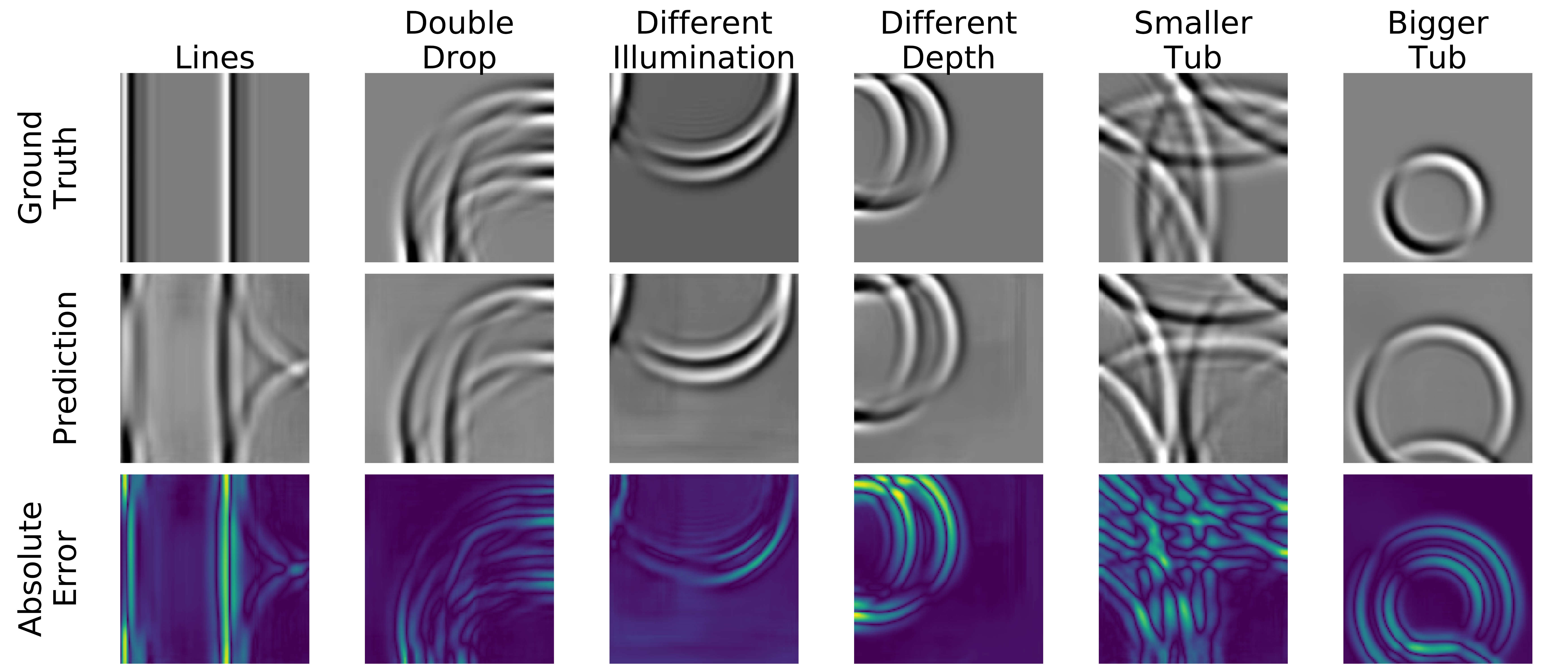}
\end{minipage}
  \caption{\textbf{Left: Qualitative comparison between the U-Net and the PredRNN++ on the test set.} The intensity profile corresponds to the yellow line (the line with the highest variation). In this particular case, the PredRNN++ has missed the time constant. \textbf{Right: Predictions (at time $t=80$) of the U-Net in the various dataset that have not been seen during training.} In double drop, we see how the model fails to accurately predict the double wave-front. For bigger and smaller tub it misses the time constant.}
  \label{fig:res:gen_qual}
  \vskip -2mm
\end{figure}

\subsection{Generalisation: Extrapolation in other physical settings}

Here, we evaluate the capabilities and limitations of our models by testing under different initial conditions, illumination models and tank dimensions (Table \ref{app:tab:dataset} in Appendix). For conciseness, we only present the results of the U-Net but the same conclusions stand for all the models.

The U-Net seems to be quite robust to changes in illumination. The RMSE for opposite illumination angle ($135^\circ$) is indistinguishable to the original test set (Figure \ref{fig:res:gen} and Table \ref{tab:res:gen}). This indicates that the learned representation is invariant to a perpendicular phase shift in lighting conditions.
Propagation of linear waves appears to be more challenging, RMSE exceeds 0.10 after just 12 frames. The visualisation shows how the morphology of the prediction is qualitatively different, containing circular artefacts, reminiscent of the training data (Figure \ref{fig:res:gen_qual}). When two drops are used, the RMSE is fairly low but the two wave-fronts of the predictions are sometimes blurred. We also varied the tank size to study the effect of wave speed. It seems that both cases are challenging with the smaller tank size, or equivalently faster waves, exceeding 0.10 RMSE after just 5 frames. Predictions in Figure \ref{fig:res:gen_qual} demonstrate how the network miscalculates the wave speed, and its predictions are either faster or slower than the ground truth. Please note that direct comparisons between datasets based on the RMSE is not without shortcomings. Each dataset has its own inherent "variation" which affect the RMSE, i.e. waves move faster in a small tank (see Figure \ref{app:fig:res:gen_baselines} in the Appendix for a discussion).

\section{Conclusions and Future Work}\label{sec:conclusions}

In this work we investigated the use of deep networks for approximating wave propagation. Using a U-Net architecture, we managed to reduce the long-term approximation RMSE to 0.071 against the previous baseline of 0.186. At the same time, the U-Net is $240\times$ faster than the simulation. Our results suggest that the U-Net outperforms state-of-the-art recurrent models.
It is unclear why U-Net models perform so well in this task. It been demonstrated that convolutional networks are effective at modelling one-dimensional temporal sequences \citep{bai2018empirical}; it might be true for higher-dimensional data. Furthermore, the simulated data are based on few-step solvers. In such a case the memory modules may not offer a significant advantage. 
Lastly, we extensively assessed how the networks generalise in unseen physical settings and pointed out current limitations.

In the future, we aim to introduce noise in the simulation so the system becomes stochastic. It would be interesting to see if in this case the recurrent models learn the dynamics better than the \mbox{U-Net}. A big shortcoming of the current models is generalisation in other physical settings. We plan to address this by a physics-inspired latent space factorisation and meta-learning.


\bibliography{references, mendeley}
\bibliographystyle{iclr2020_conference}

\appendix
\section*{Appendix}

\section{Datasets}\label{app:sec:datasets}

The datasets were created by simulating the Saint-Venant equations:
\begin{equation}\label{eq:saint-venant}
\begin{array}{r}
{h_{t}+((H+h) u)_{x}+((H+h) v)_{y}=0} \\
{u_{t}+u u_{x}+v u_{y}+g h_{x}-\nu\left(u_{x x}+u_{y y}\right)=0} \\
{v_{t}+u v_{x}+v v_{y}+g h_{y}-\nu\left(v_{x x}+v_{y y}\right)=0}
\end{array}
\end{equation}

The package triflow \citep{triflow} was used for the simulation. The Coriolis force and viscosity terms were neglected, kinematic viscosity was $10^{-6} m^2/s$ which is close to water viscosity at $20^\circ$C, the height H is set to 10 m and the size of the tank is randomly selected in each simulation between 10 and 20m. The initial wave excitation is in the form of a Gaussian droplet at random locations. For rendering, we used $45^{\circ}$ lighting azimuth and $20^{\circ}$ altitude. Each sequence is 100 steps long while the time step is 0.01 sec. In total, 3,000 sequences were rendered. The frame size was $184\times 184$ pixels but was subsequently re-sampled down to $128\times 128$. The generalisation datasets were created with the same method by varying the physical properties of the simulation (Table \ref{app:tab:dataset}).

We also used image normalisation which is known to improve performance on image prediction tasks. Normalising the pixel values to zero mean and standard deviation 1 worked best for us. Note that the normalising values are computed from the training set alone and applied to the validation and test sets. Data augmentation techniques like horizontal and vertical flips were employed on a per sequence basis. From the 3000 sequences of the original dataset, 70\% were used for training, 15\% for validation and 15\% for testing.

\begin{table}[h]
\small
\begin{center}
\begin{tabular}{llrrrr}
\hline
\textbf{Dataset Name} & \textbf{Initial Condition} & \multicolumn{1}{l}{\textbf{Height(m)}} & \multicolumn{1}{l}{\textbf{Tank Size(m)}} & \multicolumn{1}{l}{\textbf{Illum. Azimuth}} & \multicolumn{1}{l}{\textbf{Sequences}} \\ \hline
Training/Validation/Test              & Droplet                    & 10                                        & {[}10, 20{]}                           & $45^{\circ}$                                   & 3000                                   \\
Double Drop           & Double Droplet             & 10                                        & {[}10, 20{]}                           & $45^{\circ}$                                   & 500                                    \\
Lines                 & Line wave                  & 10                                        & {[}10, 20{]}                           & $45^{\circ}$                                   & 500                                    \\
Opposite Illumination  & Droplet                    & 10                                        & {[}10, 20{]}                           & $135^{\circ}$                               & 500                                    \\
Random Illumination   & Droplet                    & 10                                        & {[}10, 20{]}                           & Random                               & 500                                    \\
Shallow Depth        & Droplet                    & 5                                         & {[}10, 20{]}                           & $45^{\circ}$                                   & 500                                    \\
Small Tank          & Droplet                    & 10                                        & {[}5, 10{]}                            & $45^{\circ}$                                   & 500                                    \\
Big Tank           & Droplet                    & 10                                        & {[}20, 40{]}                           & $45^{\circ}$                                   & 500                                    \\
\end{tabular}
\end{center}
\caption{The original dataset was used for training, model selection and evaluation. Models were also trained with the fixed tank dataset to study the effect of tank size. All the other datasets were used to evaluate the generalisation capabilities of the models.}
\label{app:tab:dataset}
\end{table}

\section{Models}\label{app:sec:mod}

\subsection{LSTM}\label{app:sec:mod:lstm}

The encoder consists of 4 convolutional layers with 60, 120, 240, 480 feature maps, kernel sizes 7, 3, 3, 3 and padding of 2, 1, 1, 1 pixels. Dimensionality reduction is achieved by using a kernel stride of size $2$ in all layers. After each convolutional layer, there is a batch normalisation layer and a \textit{tanh} non-linearity. In the last convolutional layer, dropout is used on $25\%$ of the units, that are chosen randomly in each pass. The final part of the encoder is a fully connected layer of width $\mathcal{L}=1000$. This is the latent vector input to the three LSTMs. One LSTM is used for the first input, the second LSTM is for predicting the 10th frame (midway) and the third LSTM for all the other frames. The decoder is based on deconvolutions that double the spatial dimensions of the feature maps in each layer until the original $128\times128$ size is reached. It is a mirror of the encoder in terms of feature map size while the kernel is 3, the padding is 1 and the stride is 2 for all the layers. Figure \ref{app:fig:mod:lstm_architecture} depicts the architecture.

\begin{figure}[h]
\centering
  \includegraphics[width=.95\linewidth]{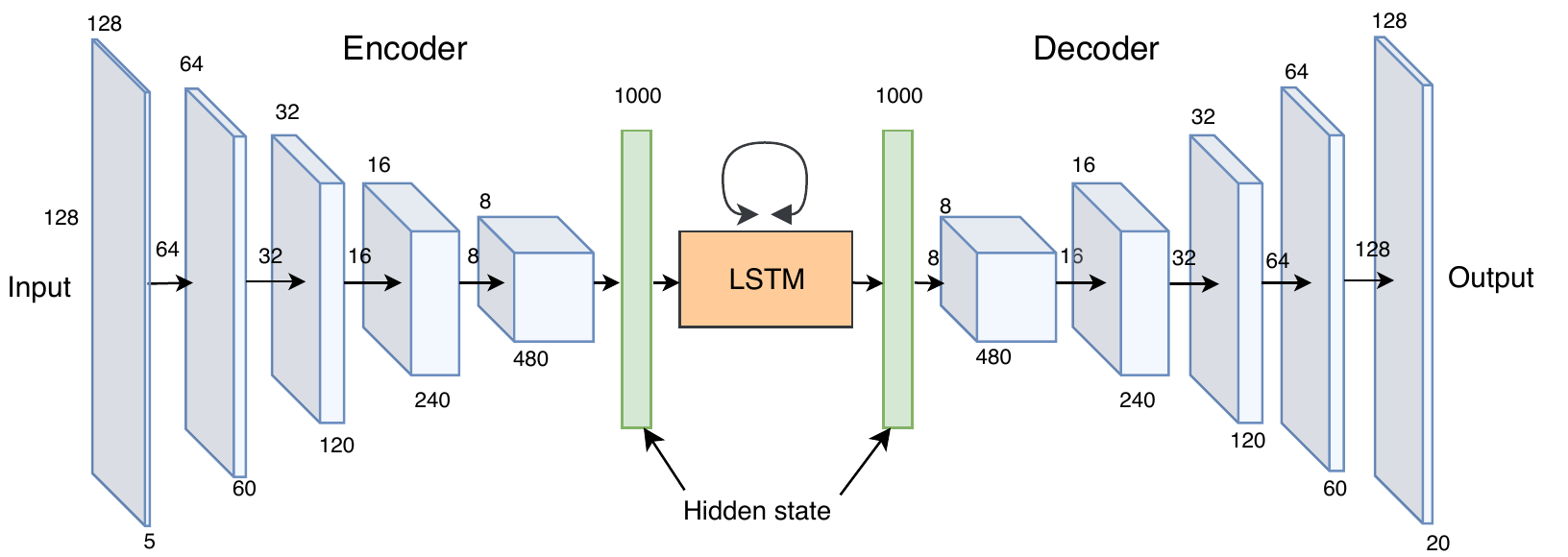}
  \caption{\textbf{Schematic of the encoder and decoder used in the LSTM model \citep{Sorteberg2019ApproximatingNetworks}. }Dimensions or layers are left and up, number of channels at the bottom.}
  \label{app:fig:mod:lstm_architecture}
\end{figure}

\begin{figure}[h]\label{app:fig:mod:lstm_train}
\centering
  \includegraphics[width=.95\linewidth]{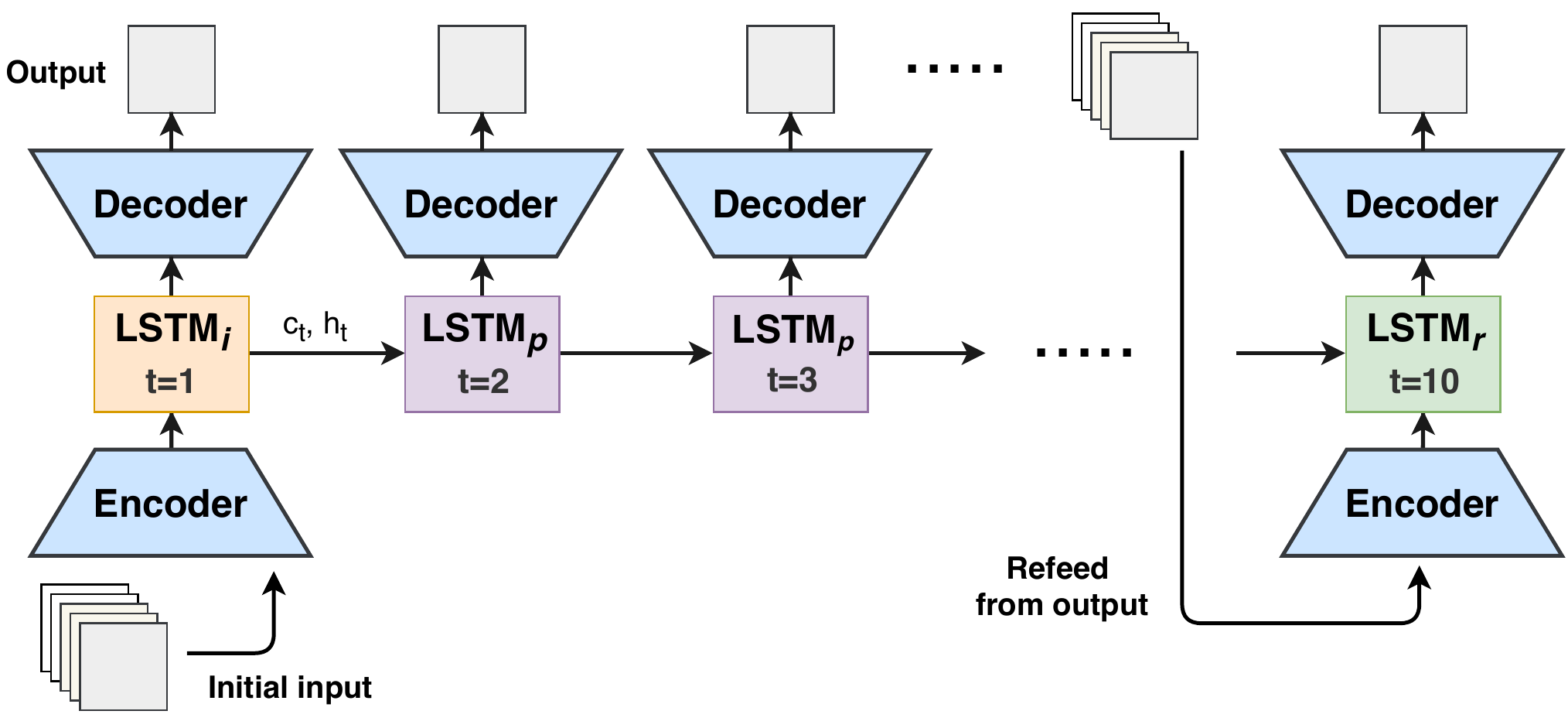}
  \caption{\textbf{Schematic of the encoder and decoder used in the LSTM model \citep{Sorteberg2019ApproximatingNetworks}. }Dimensions or layers are left and up, number of channels at the bottom.}
  \label{app:fig:mod:lstm_architecture}
\end{figure}

\subsection{ConvLSTM}\label{app:sec:mod:convlstm}

Our architecture uses a stack of 3 ConvLSTM cells. Initially, a convolutional encoder with  8, 64, 192 feature maps respectively reduces the spatial dimensions to $31 \times 31$. All layers have kernels of size 3, zero padding of width 1 and Leaky ReLU non-linearities with slope 0.2. A stride of 2 pixels is used to reduce the dimensionality. At the final layer, the input is represented by a $16\times16\times192$ tensor.  Inside the ConvLSTMs we use kernels of size 3 and zero padding of 1 pixel to avoid the dimensionality reduction. The decoder uses deconvolutions with stride 2 to double up the pixel dimensions in each layer. 

\begin{figure}[h]
  \centering
  \includegraphics[width=.95\linewidth]{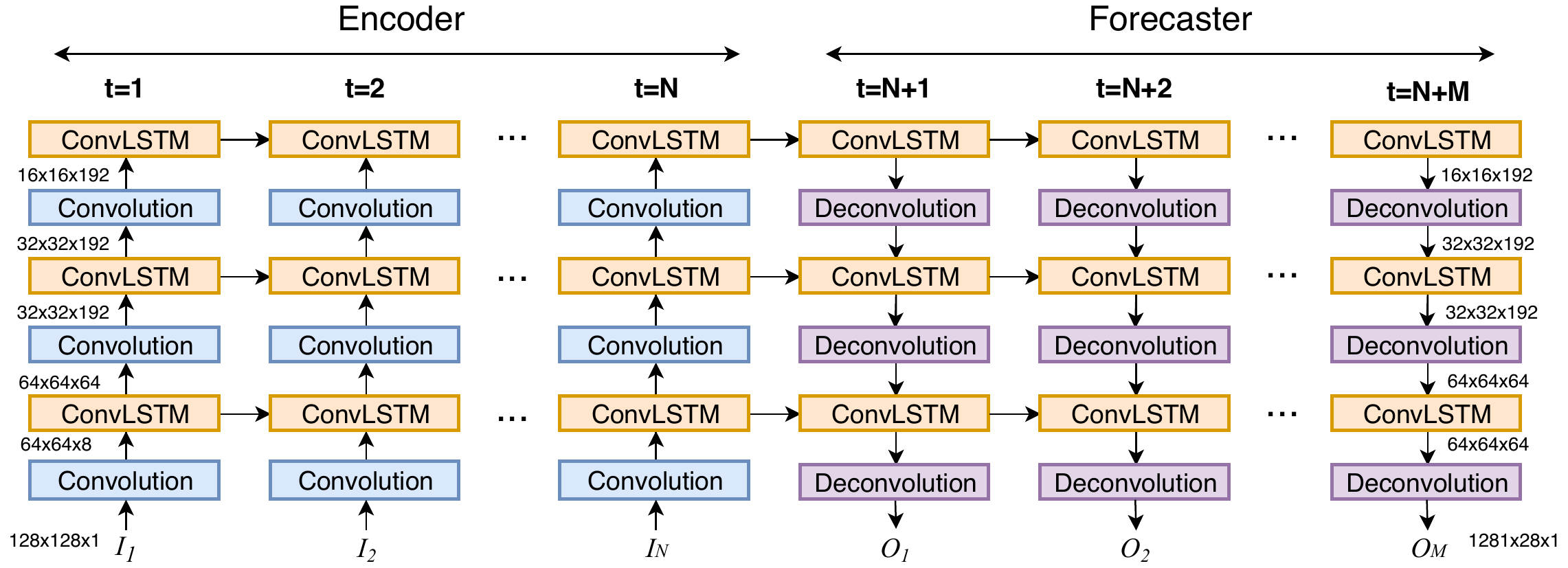}
  \caption{\textbf{ConvLSTM Model} The encoder processes the N=5 input frames one at a time to create an internal representation. The representation gets copied over to the forecaster that uses it to generate M=10 future frames. The feature map dimensions can be seen next to each layer.}
  \label{app:fig:mod:convlstm}
\end{figure}

\subsection{PredRNN++}\label{app:sec:mod:causallstm}

\begin{figure}[h]
  \centering
  \includegraphics[width=.95\linewidth]{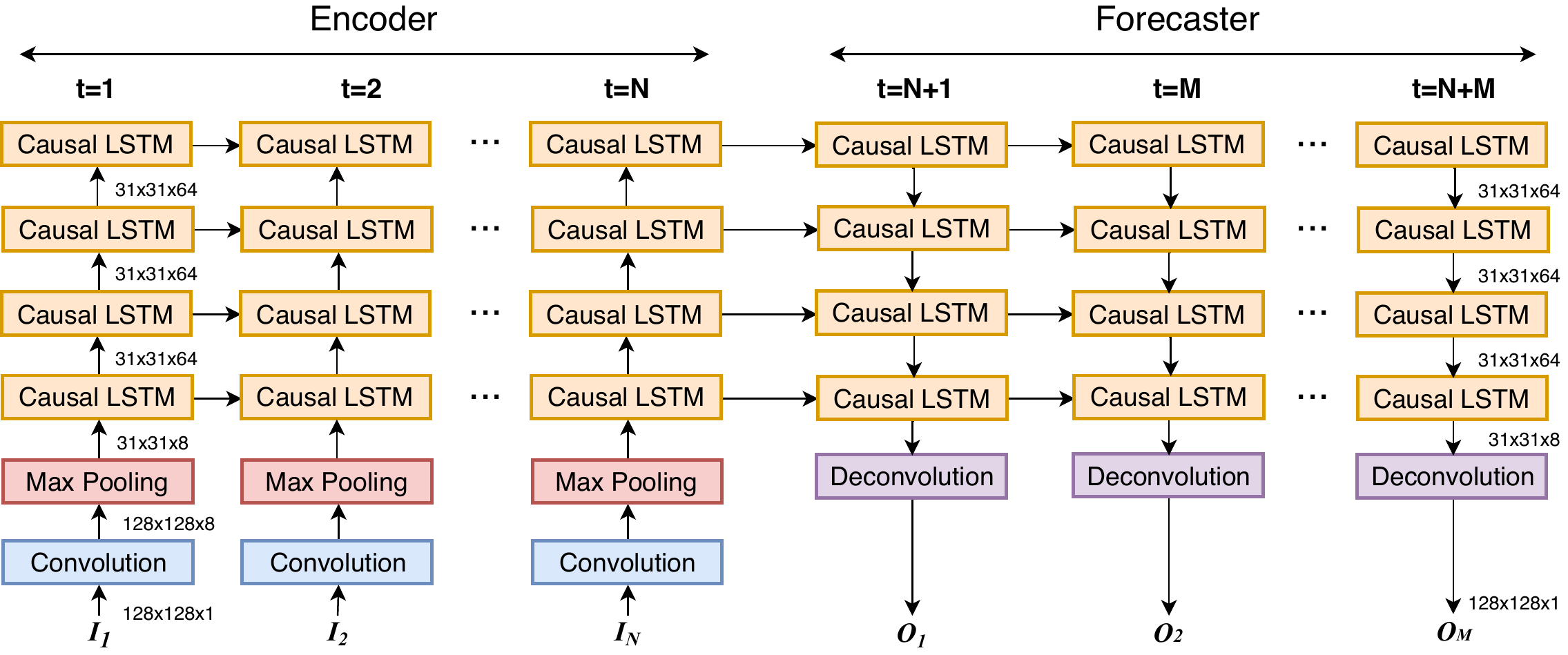}
  \caption{\textbf{PredRNN++ Model} The encoder processes the N=5 input frames one at a time and an internal representation is created in each PredRNN++ cell. These representations get copied over to the forecaster that uses them to generate M=20 future frames. The feature map dimensions can be seen next to each layer.}
  \label{app:fig:mod:causalstm}
\end{figure}

The unfolding of the model through time is presented in Figure \ref{app:fig:mod:causalstm}. The vertical stack is comprised of one convolutional, one max pooling and four PredRNN++ layers. The convolutional layer has a kernel of size 3, no padding and outputs 8 feature maps. In the original paper, they do not use any dimensionality reduction because their input dimensions are $64\times 64$ per frame. Our input dimensions ($128\times 128$) are too big to fit in available GPU memory, so we used max-pooling with stride 4 to reduce the dimensions to $31\times 31$ pixels. Following the original paper, we used 4 PredRNN++ layers but reduced the size of all of them to 64 channels each to meet hardware memory constraints. We used convolutional kernels of size 3. The forecaster uses a deconvolutional layer kernel size 7 and stride 4 to restore the internal state to the original dimensions.

\subsection{U-Net}\label{app:sec:mod:unet}

\begin{figure}[h]
\centering
  \includegraphics[width=.95\linewidth]{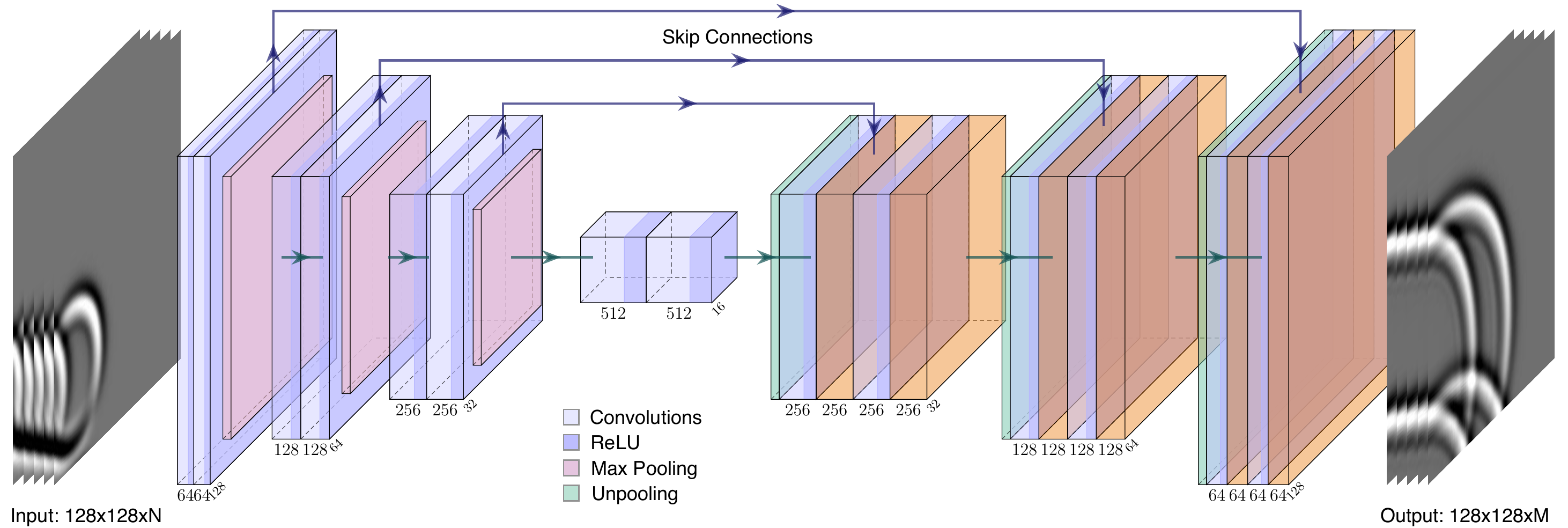}
  \caption{\textbf{Schematic of the U-Net model.} The number of channels is below each layer and its dimensions on the side. Our model has input N=5 and output M=20.}
  \label{app:fig:mod:unet}
\end{figure}

The encoder is composed of fours blocks each containing two convolutional layers with kernel size 3 and padding 1, followed by ReLU non-linearities. The first three blocks include a max-pooling layer of stride 2 that reduces the size in half. The number of feature maps doubles in each layer. For the expanding part, we use bilinear interpolation with scale factor 2 instead of deconvolutions to keep the number of parameters low. Skip connections are also employed to copy feature maps from earlier layers but contrary to the original paper we do not reduce the dimensions of the copied feature maps. This way, high-level, coarser feature maps are combined with fine-grained local information of lower layers over the whole domain. The network architecture can be seen in Figure \ref{app:fig:mod:unet}.

\section{Hyperparameters}\label{app:sec:hyperparameters}

Assume that $N$ is the number of input and $M$ the number of output frames of the model. For each training iteration, we randomly selecting $K$ sub-sequences of length $N+M$ from each simulated sequence. The models were trained to minimise the MSE over their respective output length $M$. In each iteration, the weights are updated using an Adam optimiser while a scheduling scheme adjusts the learning rate (LR) by a scaling factor of $10^{-2}$ if there is no improvement in validation error after a given amount of epochs (patience). The hyperparameters of interest are the input length $N\in \{3,5,10\}$, training output length $M\in \{1,5,10,20\}$, samples per sequence between weight updates $K\in \{1, 3, 5, 10\}$, batch size $b\in \{16, 32\}$, LR $\in \{10^{-2}, 10^{-3}, 10^{-4}, 10^{-5}\}$ and patience $p\in \{5,7\}$. Grid search was used to find the best set of hyperparameters of each model. The training budget was 24h hours. To obtain an arbitrary long prediction we the output as the next input. The goal is to obtain networks with a low error in long term prediction so, during model selection, we chose the hyper-parameters that gave the lowest validation error over 50 frames regardless of the output size of the model $M$. The final hyperparameters and model sizes can be found in Tables \ref{app:tab:hp} and  \ref{app:tab:train}.

\begin{table}[h]
\centering
\begin{tabular}{lcccccc}
\toprule
\textbf{Model} & \textbf{\begin{tabular}[c]{@{}c@{}}Input\\ Length\end{tabular}} & \textbf{\begin{tabular}[c]{@{}c@{}}Output\\ Length\end{tabular}} & \textbf{\begin{tabular}[c]{@{}c@{}}Samples per\\ Sequence\end{tabular}} & \textbf{\begin{tabular}[c]{@{}c@{}}Batch \\ Size\end{tabular}} & \textbf{\begin{tabular}[c]{@{}c@{}}Learning\\ Rate\end{tabular}} & \textbf{Patience} \\
\hline
LSTM           & 5                                                               & 20                                                               & 10                                                                      & 16                                                             & $10^{-4}$                                           & 5                 \\
ConvLSTM       & 5                                                               & 10                                                               & 5                                                                       & 8                                                              & $10^{-3}$                                           & 7                 \\
PredRNN++    & 5                                                               & 20                                                               & 5                                                                       & 4                                                              & $10^{-4}$                                           & 3                 \\
U-Net           & 5                                                               & 20                                                               & 10                                                                      & 16                                                             & $10^{-4}$                                           & 7    \\
\bottomrule
\end{tabular}
\caption{Hyper-parameters of the best performing model for each architecture}
\label{app:tab:hp}
\end{table}

\newpage
\section{Model size and speed}

Models were implemented in PyTorch and the code is publicly available in GitHub. Models were trained on a GTX 1060 GPU with 6GB of memory. Total training time includes evaluation overhead. 

\begin{table}[h]
\centering
\begin{tabular}{lccccc}
\toprule
\textbf{Model} & \textbf{\begin{tabular}[c]{@{}c@{}}Trainable\\ Parameters\end{tabular}} & \textbf{\begin{tabular}[c]{@{}c@{}}Epoch \\ Time\end{tabular}} & \textbf{\begin{tabular}[c]{@{}c@{}}Num.\\ Epochs\end{tabular}} & \textbf{\begin{tabular}[c]{@{}c@{}}Best\\ Epoch\end{tabular}} & \textbf{\begin{tabular}[c]{@{}c@{}}Total\\ Training Time\end{tabular}} \\
\hline
LSTM           & 88.2M                                                                   & 12m                                                            & 75                                                            & 71                                                            & 24h                                                                    \\
ConvLSTM       & 12.3M                                                                   & 36m                                                            & 24                                                            & 18                                                            & 24h                                                                    \\
PredRNN++    & 2.5M                                                                    & 33m                                                            & 43                                                            & 36                                                            & 24h                                                                    \\
U-Net           & 7.8M                                                                    & 8m                                                             & 171                                                           & 166                                                           & 24h\\
\bottomrule
\end{tabular}
\caption{Model size and training times}
\label{app:tab:train}
\end{table}

\begin{table}[h]
\begin{tabular}{lrr}
\toprule  
Method  &   Time per frame (ms) & Speed-up\\
\midrule
Numerical simulator  &  630.7 & - \\
LSTM & 15.0 & 40x \\
ConvLSTM & 4.5 & 141x \\
PredRNN++ & 9.2 & 68x \\
U-Net & 2.6 & 241x \\
\bottomrule
\end{tabular}
\caption{\textbf{Time it takes to compute one frame.} Deep learning approximations offer a significant speed-up over numerical simulations.}
\label{app:tab:speed}
\end{table}

\newpage

\section{Results Addendum}

\begin{figure}[h]
\centering
\begin{minipage}[c]{.5\textwidth}
  \centering
  \includegraphics[width=\linewidth]{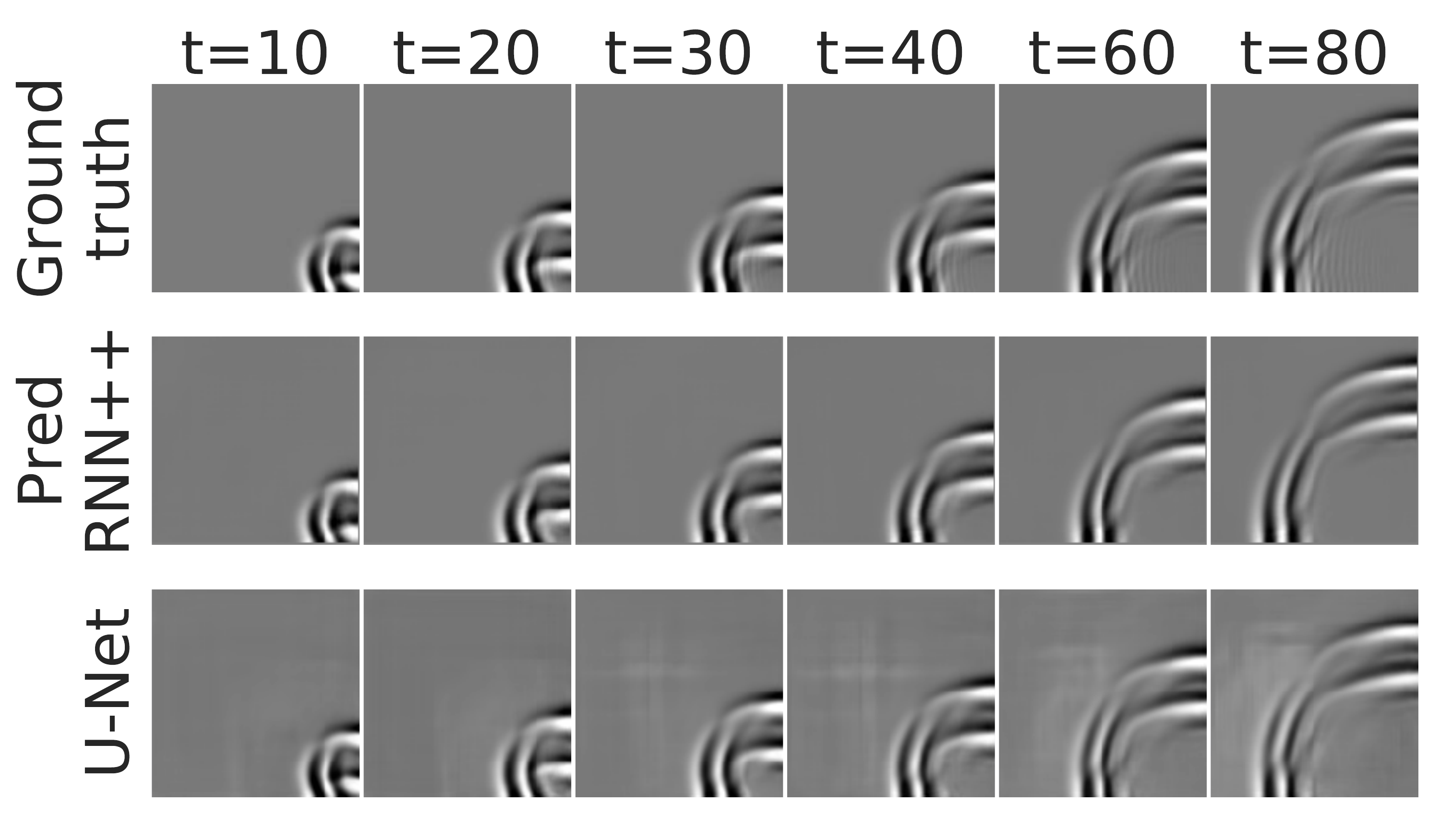} 
\end{minipage}%
\begin{minipage}[c]{.5\textwidth}
  \centering
  \includegraphics[width=\linewidth]{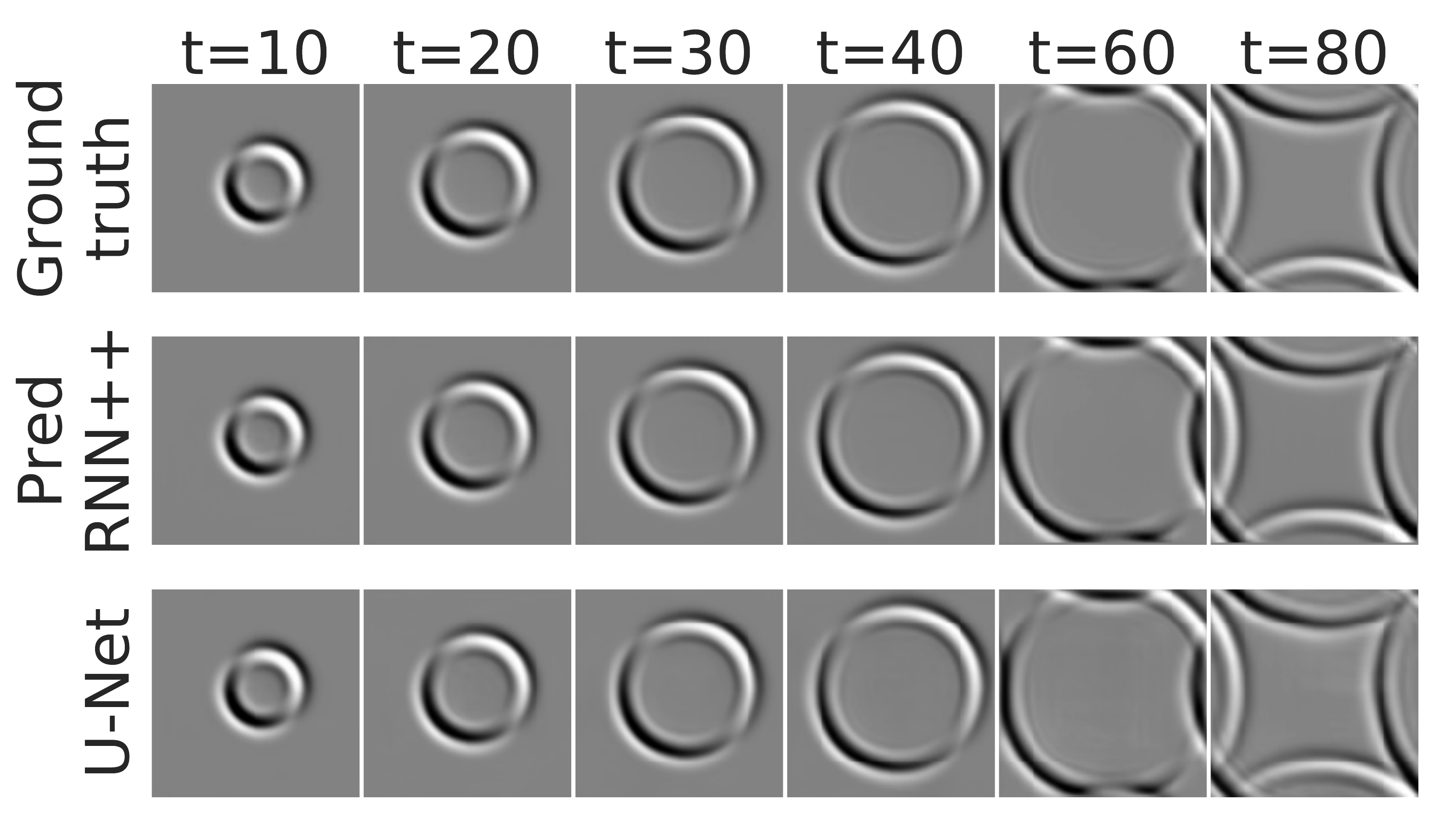}
\end{minipage}
  \caption{Prediction roll-out from U-Net and PredRNN++. Both sequences are from the test set.}
  \label{app:fig:res:timeseq}
\end{figure}

\begin{figure}[h]
\centering
\begin{minipage}[c]{.5\textwidth}
  \centering
  \includegraphics[width=\linewidth]{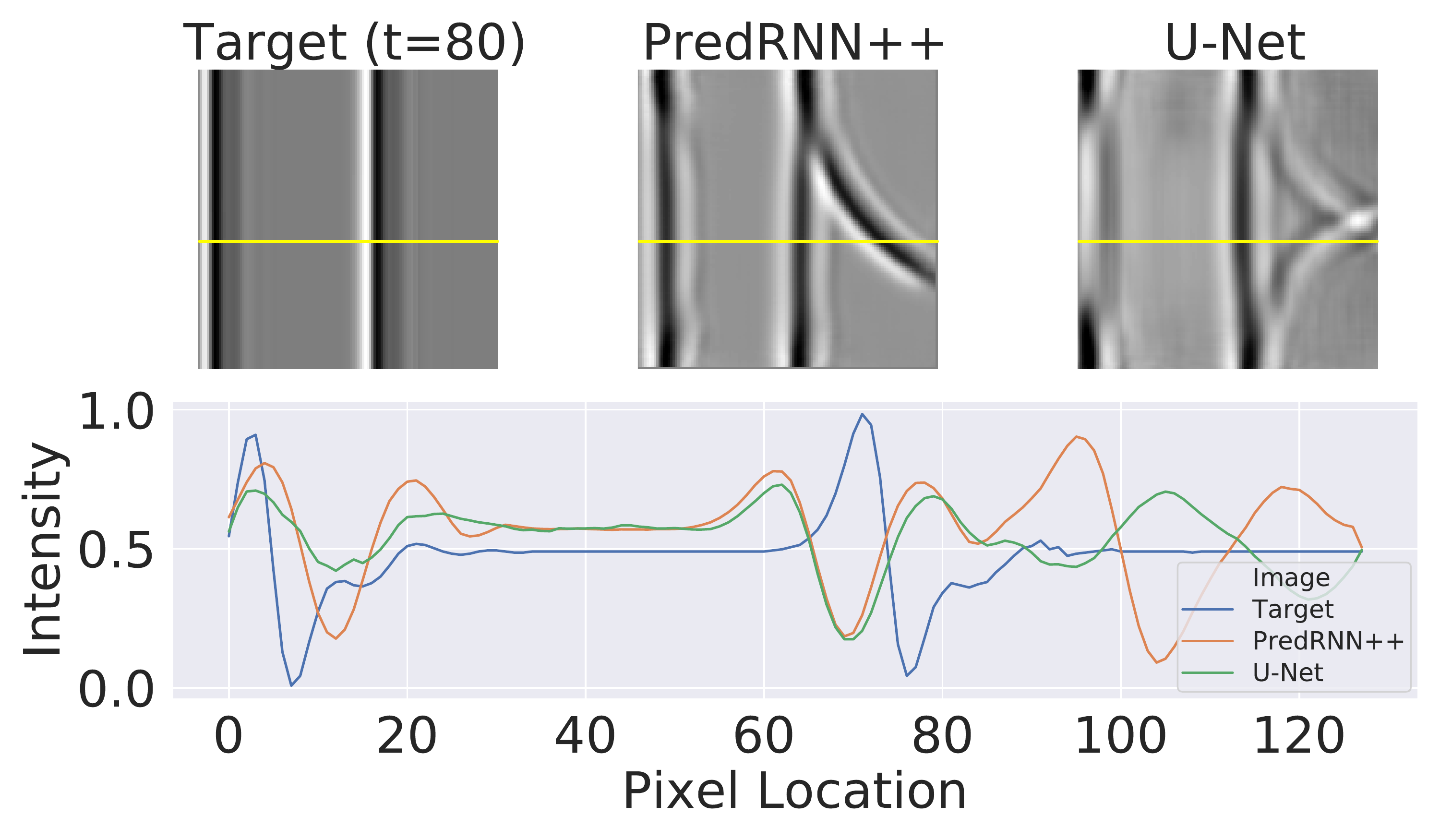} 
\end{minipage}%
\begin{minipage}[c]{.5\textwidth}
  \centering
  \includegraphics[width=\linewidth]{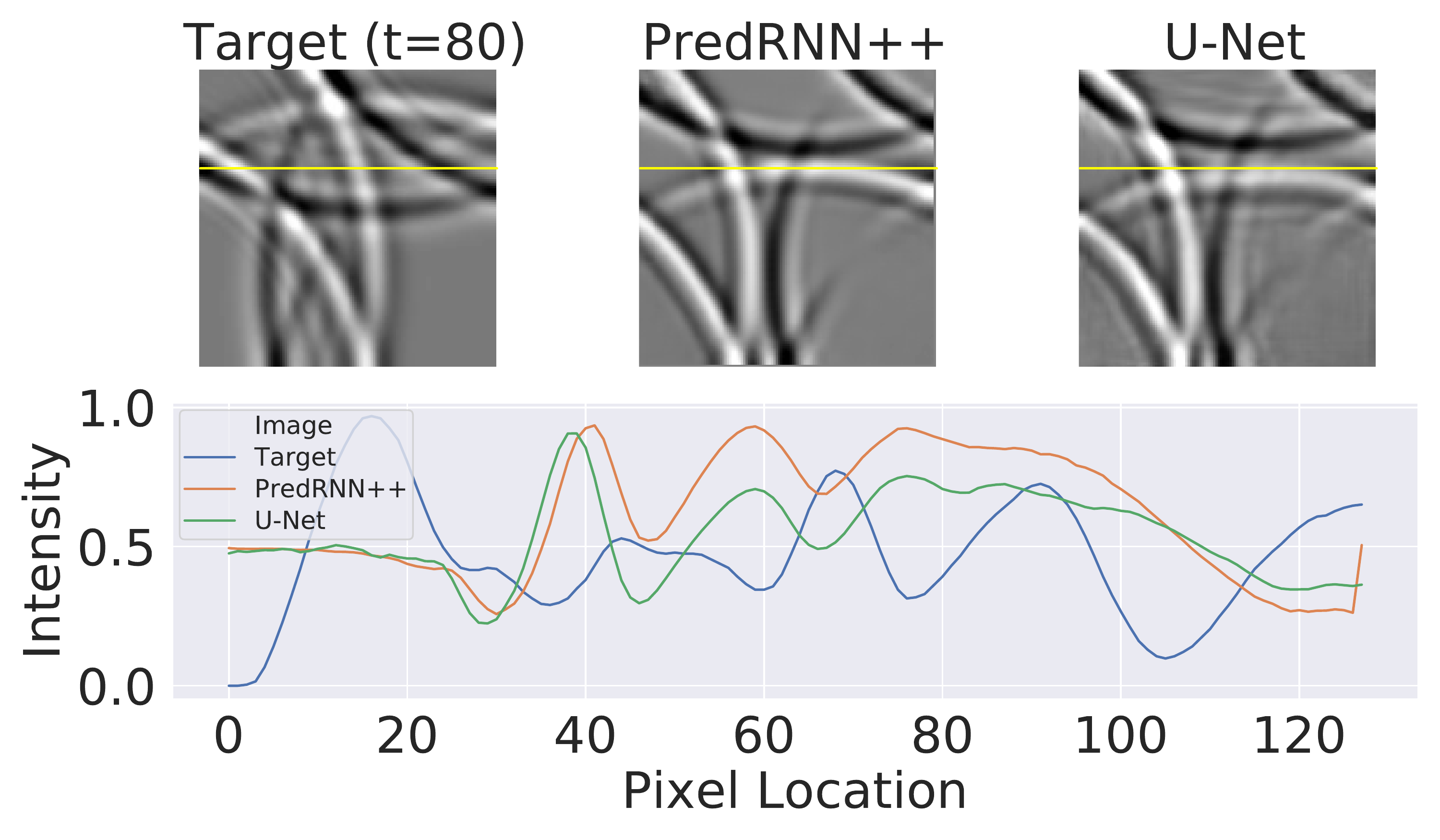}
\end{minipage}
  \caption{Generalization of U-Net and PredRNN++ in different physical settings. In the left we see linear waves. The networks introduce circular patterns where they don't exist. In the right panel it is the smaller tank, where waves are faster. Both models miss the time constant by being slower than the ground-truth.}
  \label{app:fig:res:gen_baselines}
\end{figure}

\begin{figure}[ht]
\centering
\begin{minipage}[c]{.5\textwidth}
  \centering
  \includegraphics[width=\linewidth]{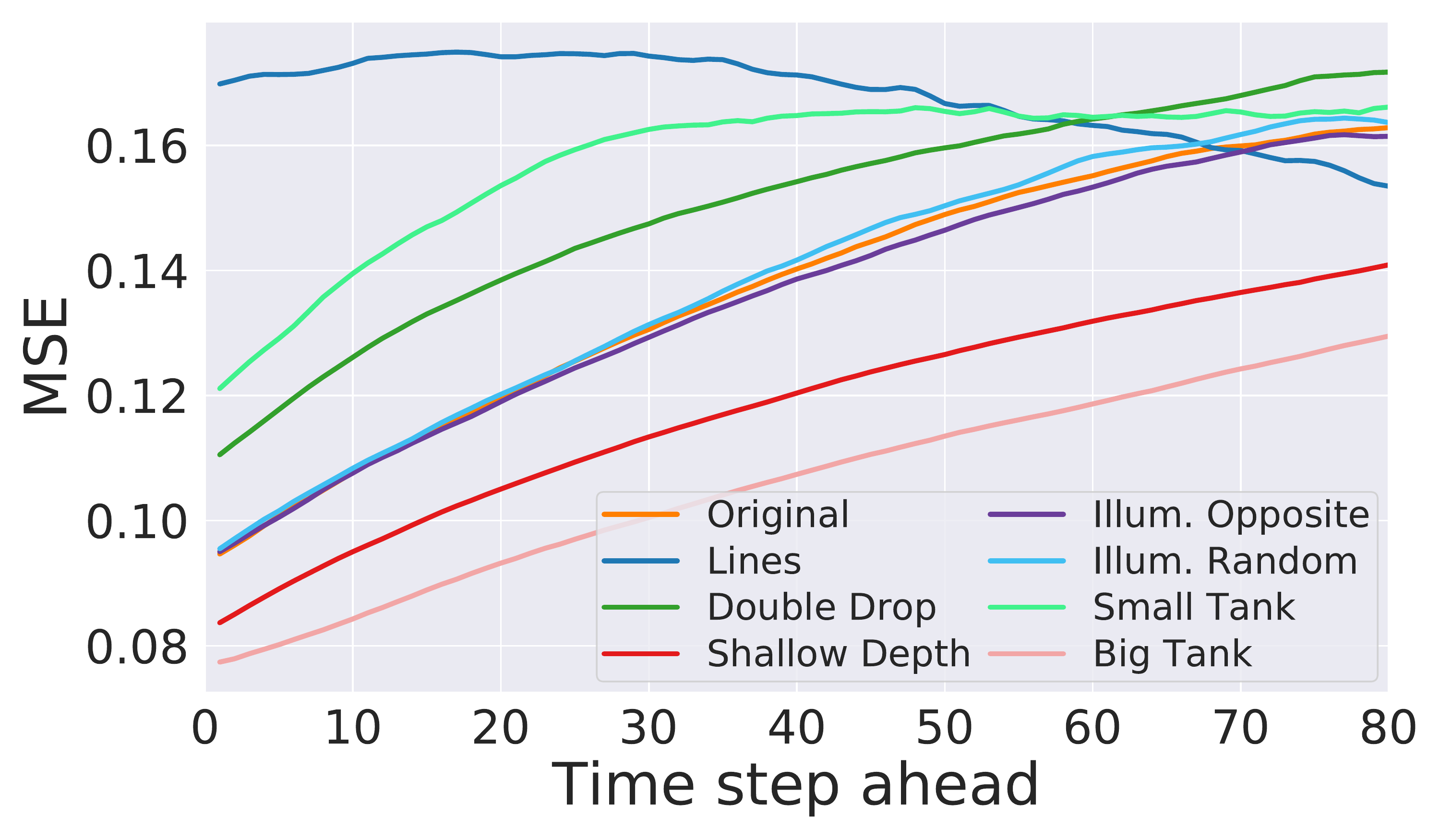} 
\end{minipage}%
\begin{minipage}[c]{.5\textwidth}
  \centering
  \includegraphics[width=\linewidth]{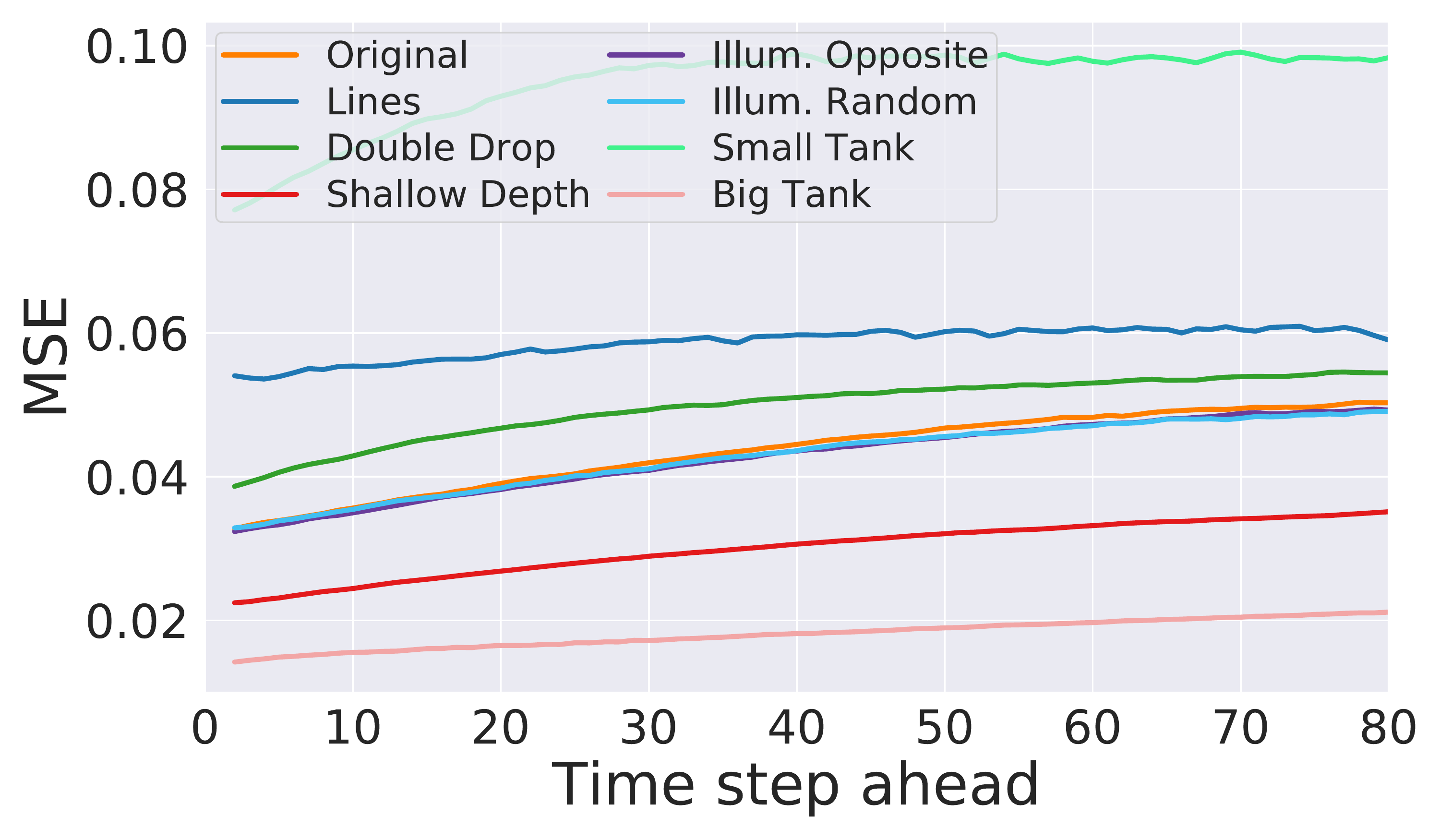}
\end{minipage}
  \caption{\textbf{Flat Image (left)} and \textbf{Previous Frame (right)} baselines compared against ground truth across different datasets. The flat image baseline measures how much the frame is away from the reference height. The Previous Frame indicates how much consecutive frames differ. Datasets are not equally hard to predict and this should be taken into account when assessing the generalisation capacity of a model.
  }
  \label{add:fig:res:gen_baselines}
\end{figure}

\subsection{Predicting the tank size from the latent space}

Here we check if the trained U-Net acquired any understanding of the physical properties of the system. We focus on the tank size, or inversely the speed of the wave, for two reasons. First of all, the U-Net failed to extrapolate to different tank sizes. This experiment could provide some insights on why this failure happens. Secondly, tank size information is readily available. Each dataset sequence corresponds to a different tank size, and the tank is always square. In the training and testing dataset we have tank size $s_i \in [10,20]$ meters. For the smaller tank we used $s_i \in [5,10]$ and for the bigger $s_i \in [20,40]$ meters. 

\begin{figure}[h]
\centering
  \includegraphics[width=.8\linewidth]{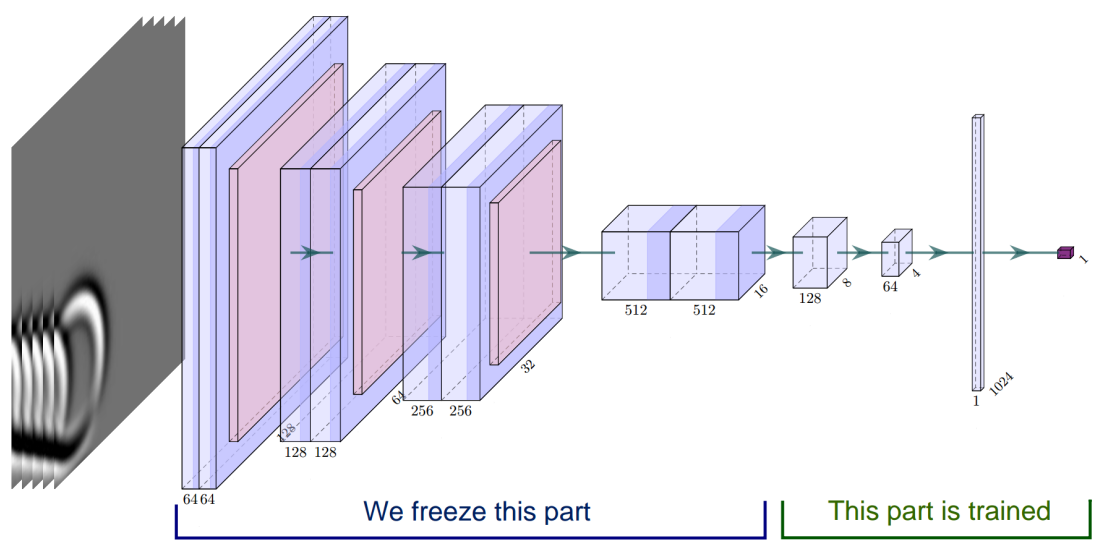}
  \caption{\textbf{Schematic of the model used for tank size prediction}}
  \label{app:fig:mod:unet-tank}
\end{figure}

The question we try to answer is: does the latent representation of the U-Net capture that tank size information $s_i$. We take the pre-trained encoder from the U-Net and add some additional layers so that the output is only one number (Figure \ref{app:fig:mod:unet-tank}). The system is trained to predict the tank size when given 5 consecutive frame. Only the additional part is updated during training. The weights of the encoder are kept frozen. We compare the pre-trained encoder against a randomly initialised encoder. We, also, compare the models to a dummy regressor that predicts always the mean tank size for each dataset i.e. 15 for the test set, 7.5 for the small tank and 30 for the big tank. Results in Table \ref{app:tab:speed} indicate that the pre-trained encoder can be used to extract the tank size with relatively low error (0.14) while the random encoder gives a much higher error of 2.27, slightly lower to the dummy regressor (2.45). This indicates that the pre-trained encoder encapsulates physically relevant information relating to the tank size. When it comes to the bigger and smaller tanks, both the pre-trained and the random encoders fail to extrapolate and give errors higher than the dummy regressor. 

\begin{table}[h]
\begin{tabular}{lrrr}
\toprule  
  &   Test set & Bigger Tank & Smaller Tank\\
\midrule
Pre-trained encoder  & \textbf{0.14} & 6.65 & 2.23 \\
Random encoder & 2.27 & 14.21 & 6.35  \\
Dummy regressor & 2.45 &\textbf{ 5.19} &\textbf{ 1.22} \\
\bottomrule
\end{tabular}
\caption{\textbf{Predicting the tank size from the U-Nets latent space.} The pre-trained model learns to identify the correct tank size with a relatively low error. The randomly initialised encoder fails to do so, only marginally improving the error over a dummy regressor. A dummy regressor is one that always predict the mean tank size of the dataset. Both the pre-trained and random encoder fail to extrapolate to the smaller and bigger tanks.
}
\label{app:tab:tanksize}
\end{table}

\end{document}